\documentclass[10pt,twocolumn,letterpaper]{article}

\usepackage{iccv}
\usepackage{times}
\usepackage{epsfig}
\usepackage{graphicx}
\usepackage{amsmath}
\usepackage{amssymb}
\usepackage{subfigure}
\usepackage{bbm}
\usepackage{autobreak}
\allowdisplaybreaks
\usepackage[pagebackref=true,breaklinks=true,letterpaper=true,colorlinks,bookmarks=false]{hyperref}

\DeclareMathOperator*{\E}{\mathbb{E}}

\DeclareMathOperator*{\argmin}{arg\,min}

\iccvfinalcopy % *** Uncomment this line for the final submission

\begin{document}

%%%%%%%%% TITLE
\title{CA-UDA: Class-Aware Unsupervised Domain Adaptation with \\ Optimal Assignment and Pseudo-Label Refinement}
%%  Domain Adaptation with Progressive Target Label-Oriented Refinement
%%Uncertainty-Guided Target Label Correction
\author{Can Zhang\\
National University of Singapore\\
{\tt\small can.zhang@u.nus.edu}
\and
Gim Hee Lee\\
National University of Singapore\\
{\tt\small dcslgh@nus.edu.sg}
}

\maketitle
\ificcvfinal\thispagestyle{empty}\fi

%%%%%%%%% ABSTRACT
\begin{abstract}
Recent works on unsupervised domain adaptation (UDA) focus on the selection of good pseudo-labels as surrogates for the missing labels in the target data. 
However, source domain bias that deteriorates the pseudo-labels can still exist since the shared network of the source and target domains are typically used for the pseudo-label selections. The suboptimal feature space source-to-target domain alignment can also result in unsatisfactory performance. 
In this paper, we propose CA-UDA to improve the quality of the pseudo-labels and UDA results with optimal assignment, a pseudo-label refinement strategy and class-aware domain alignment. We use an auxiliary network to mitigate the source domain bias for pseudo-label refinement. Our intuition is that the underlying semantics in the target domain can be fully exploited to help refine the pseudo-labels that are inferred from the source features under domain shift. 
Furthermore, our optimal assignment can optimally align features in the source-to-target domains and our class-aware domain alignment can simultaneously close the domain gap while preserving the classification decision boundaries. 
Extensive experiments on several benchmark datasets show that our method can achieve state-of-the-art performance in the image classification task.
\end{abstract}

%%%%%%%%% BODY TEXT
\section{Introduction}

Domain shift~\cite{44domainshift} refers to the phenomenon where the distribution between two data for a closely related task differs significantly. Despite the success of deep learning in many computer vision-related tasks~\cite{1resnet50, 6vgg, 36densenet, 37Alexnet}, domain shift can cause a deep network to fail catastrophically when it is trained and used on data for the same task but from different distributions. A naive approach is to augment training data across different domains~\cite{25DA1,26DA2,23ADDA}, but unfortunately, this approach quickly becomes impractical due to the expensive cost and laborious effort incurred to label large amounts of training data. Unsupervised domain adaptation (UDA) is a more pragmatic alternative, where the goal is to transfer task-specific knowledge, e.g. image classification, from a labeled source domain to an unlabeled target domain. The solution to this challenging problem can potentially be used in many practical applications such as self-driving cars, augmented/virtual reality, etc; where the labeled source data can be collected in a controlled laboratory environment, e.g. simulations, while the unlabeled target data is from a closer-to-deployment environment, e.g. parking garages and highways.  
%------------------------------------------------------------------------
\begin{figure}[t]
\centering
\noindent\makebox[0.45\textwidth][c]{\includegraphics[scale=0.35]{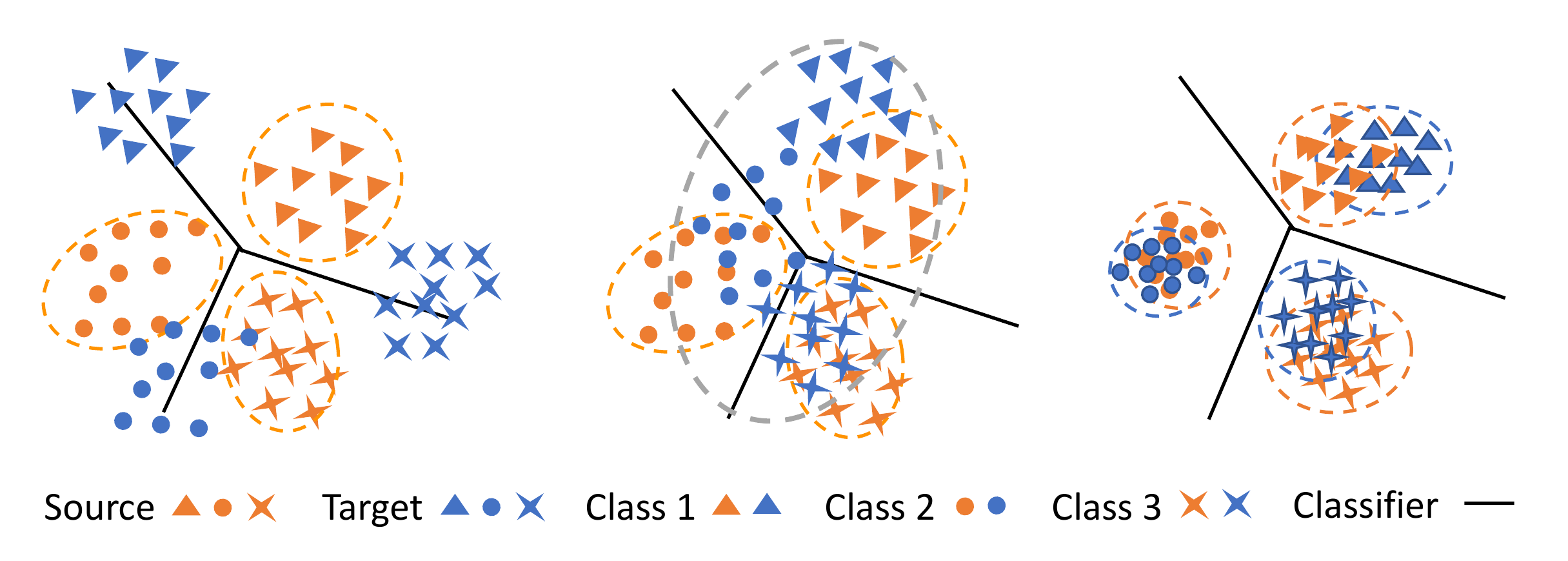}}
\caption{(Best viewed in color.) Illustration of UDA. (Left) Source-only learning from labeled source data, and (Middle) global alignment that matches marginal data distribution across domains without class information lead to poor classification results on the target data. (Right) Our class-aware alignment reduces conditional distribution discrepancy.} %\vspace{-3mm}
\label{fig:teaser}
\end{figure}

%-------------------------------------------------------------------------
% Over the years, many works~\cite{12RevGrad,9DAN,23ADDA,28CondDist2,8CAN} have been proposed to solve the UDA problem by learning domain-invariant feature representations using a shared feature extractor to align the source and target domains. These works are largely inspired by the theoretical results from~\cite{30Therory,46theory,47theory}, which state that the target error is bounded by the source error and the divergence between marginal distributions in source and target domains. Consequently, these domain-invariant feature learning works aim to minimize the source error and the source-target discrepancy via task-specific and distribution alignment losses. However, some of these works focus on aligning domain-level distributions without too much consideration on the categorical information in the target domain. As illustrated in Figure~\ref{fig:teaser} (Middle), this leads to incorrect predictions of the classes in the target domain near the class decision boundaries.

Over the years, many works~\cite{12RevGrad,9DAN,23ADDA,28CondDist2,8CAN} have been proposed to solve the UDA problem by learning domain-invariant feature representations using a shared feature extractor to align the source and target domains. These works are largely inspired by the theoretical results from~\cite{30Therory,46theory,47theory}, which state that the target error is bounded by the source error and the divergence between marginal distributions in source and target domains. Consequently, these domain-invariant feature learning works aim to minimize the source error 
% with the labeled source data 
and the source-target discrepancy via task-specific and
% additionally with unlabeled target data 
distribution alignment losses.
However, some of these works focus on aligning domain-level distributions without too much consideration on the categorical information in the target domain. As illustrated in Figure~\ref{fig:teaser} (Middle), this leads to incorrect predictions of the classes in the target domain near the class decision boundaries.

In view of the challenge in the missing categorical information from the target domain, many recent works on UDA have proposed the use of pseudo-labels as surrogates. 
%Using pseudo-labels has become a trend in recent UDA, as they compensate for the lack of label distribution in the target domain, 
% Some works ~\cite{8CAN,51pan2019transferrable,28CondDist2} perform class-conditioned alignment by introducing a loss to minimize the distance between prototypes in the source and target domain, e.g. class-level domain discrepancy loss in~\cite{51pan2019transferrable}. 
% The main limitation of these methods is the vulnerability to errors in pseudo-labels,
% % accumulation 
% as they explicitly optimize the network based on pseudo-labeled samples. 
Several works ~\cite{8CAN,24TPN,pinheiro2018unsupervised} 
directly use target domain outputs generated from the network trained by the source domain as pseudo-labels. These target domain pseudo-labels are then used together with the source domain ground truth labels to provide class-level supervision to the network while minimizing the domain discrepancy loss.   
%explicitly use pseudo-labels as supervision by performing class-conditioned alignment between prototypes in the source and target domains, e.g. class-level domain discrepancy loss in~\cite{51pan2019transferrable}. 
However, as mentioned in \cite{49progressive}, the learned classifier on source domain data might not be able to accurately predict the target samples when the domain shift is large. The accumulated errors from the erroneous pseudo-labels eventually lead to unsatisfactory results.
%The errors from pseudo-label predictions can gradually accumulate, which leads to suboptimal solution to the training procedure. 
% These methods, however, are vulnerable to errors in pseudo-labels.
% A portion of target samples lie near margins and are easily misclassified. Errors from pseudo-label predictions can gradually accumulate, which leads to suboptimal solution to the training procedure. 
Other works proposed the use of classification confidence~\cite{49progressive,27CondDist1, 28CondDist2} to select good pseudo-labels for training. Although these works have shown better results with their pseudo-label selection strategies, %they do not aim at improving the quality of all pseudo-labels in the target domain. 
source domain bias can still exist since the pseudo-label selections are done with the shared network of the source and target domains. This consequently limits the accuracy of the pseudo-labels, especially in the presence of a large domain shift.  
Furthermore, the greedy nearest-neighbor feature space source-to-target domain alignment strategy used by most existing works~\cite{49progressive,56saito2017asymmetric,27CondDist1,28CondDist2} can lead to: 1) suboptimal alignment in the feature space, and 2) poor alignment in the output classification space. 

In this paper, we propose CA-UDA to solve the UDA problem. 
Our CA-UDA improves the quality of the pseudo-labels in the target domain with
a \textbf{pseudo-label refinement} strategy. 
We train an \textit{auxiliary network} on the pseudo-labeled target samples with the easy-to-hard learning strategy. 
The source domain bias is mitigated since the auxiliary network is trained only on the target domain.
Our intuition is that the underlying semantics in the target domain can be fully exploited to help refine the pseudo-labels that are inferred from the source features under domain shift. 
We then select easy samples with confidences higher than a threshold. The confidence is predicted by the optimized target-specific network to avoid domain bias from the source data.
In addition, we introduce an \textbf{optimal assignment} step to optimally align the features of the source and target domains, and a 
\textbf{class-aware domain alignment} to simultaneously close the domain gap while preserving the classification decision boundaries using an inter-cluster and intra-cluster loss in feature and label space.

Our contributions are summarized as follows:
\begin{itemize}
    % \vspace{-2mm}
    \item We formulate a novel pseudo-label estimation and refinement procedure for UDA using optimal assignment and a self-paced easy-to-hard learning strategy.
    % \vspace{-2mm}
    \item We introduce a cross-domain center-to-center inter-cluster and an in-domain probability-to-probability intra-cluster loss to jointly minimize the class-aware discrepancy across domains (Figure~\ref{fig:teaser} (Right)). 
    % \vspace{-2mm}
    \item Our method achieves the state-of-the-art performance on 4 benchmark datasets in the classification tasks.
\end{itemize}

%-------------------------------------------------------------------------
\section{Related Work}
\label{related_work}

The work~\cite{30Therory} theoretically proves that the minimization of marginal distribution discrepancy between the source and target domains can help reduce the error in target label estimation. This work established the foundation for many subsequent works on UDA~\cite{14DANN,9DAN,23ADDA,16MCD}. 
Several works~\cite{14DANN,13MADA,12RevGrad} adopt adversarial training to learn domain-invariant feature representations. DANN~\cite{14DANN} leverages a gradient reversal layer to reduce the domain gap in feature space. ADDA~\cite{23ADDA} utilizes domain discriminator to adversarially learn the target encoder.
Other works minimize distribution divergence to align the data. These methods include: maximum mean discrepancy (MMD) in~\cite{9DAN} and correlation alignment (CORAL) in~\cite{58sun2016return}. However,
% these methods only align domain-level distribution but ignore categorical information in the target domain. As discriminability cannot be ensured in domain-invariant features, features from different classes can be mapped nearby leading to misclassification. In contrast, our method utilizes pseudo-labels to optimize the class-aware distribution alignment based on the discrepancy minimization.
these methods only align domain-level distribution.
% but ignore categorical information in the target domain.
Target samples that lie near margins of the clusters or far from their corresponding class centroids are still susceptible to mis-classification.
In contrast, our method is designed to mitigate the class-conditioned distribution shift by class-aware domain alignment.

%-------------------------------------------------------------------------

\begin{figure*}[t]
\centering
\noindent\makebox[1\textwidth][c]{\includegraphics[scale=0.6]{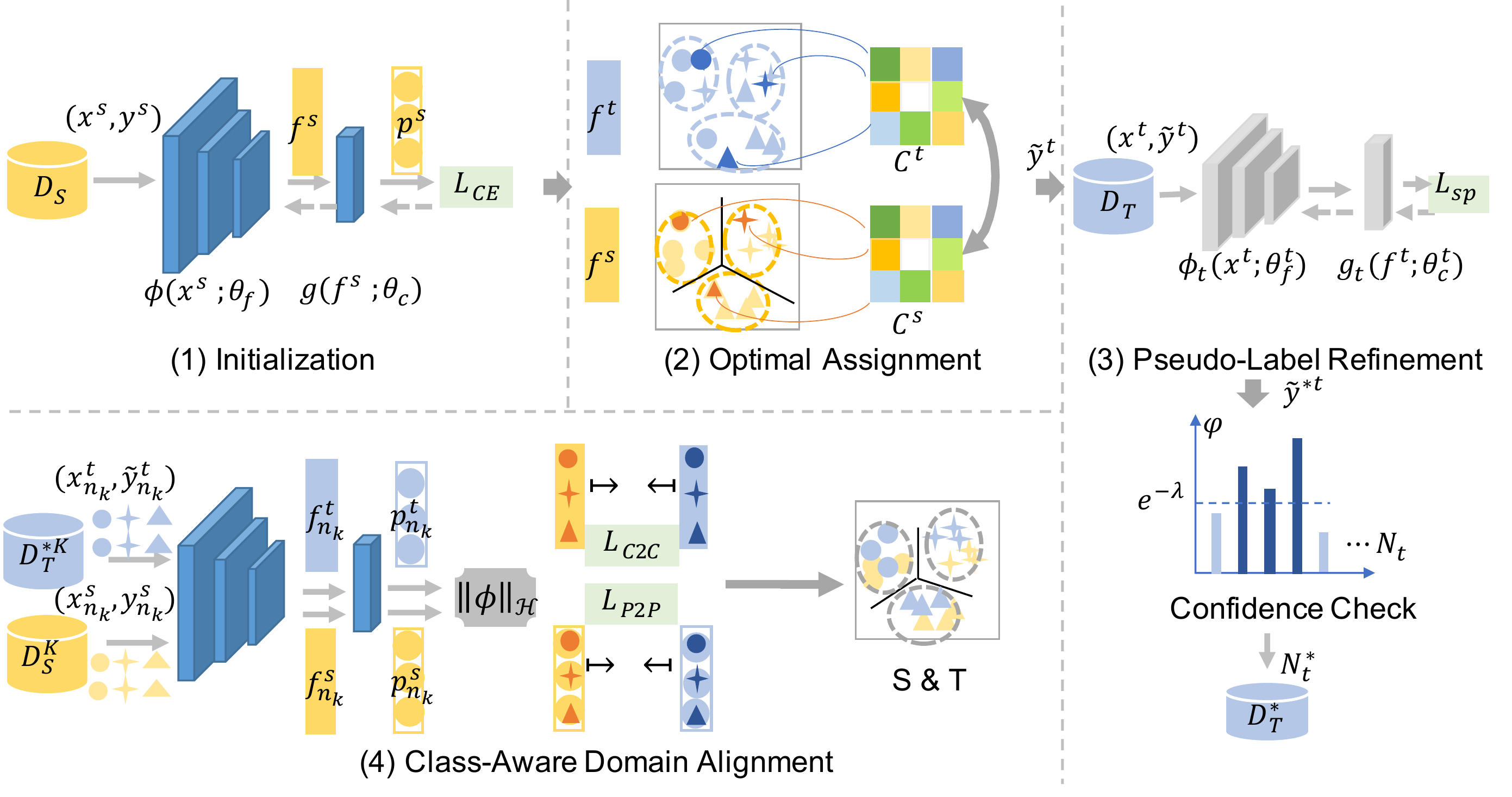}}
\caption{The overall framework of CA-UDA with four steps. 1) The network is pre-trained with the labeled source data $D_S$.
% which is expected to embed all data to the domain-invariant clustering features. 
2) Do clustering on the target feature embeddings $f^t$ to obtain cluster centroids $C^t$, and then perform the optimal assignment to generate pseudo-labels $\tilde{y}^t$ with source cluster centroids $C^s$.
% According to the similar semantic content, source and target prototypes are computed respectively, based on which, the optimal assignment is performed to generate pseudo-labels. 
3) An auxiliary network $g_t(\phi_t(x^t;\theta^t_f);\theta^t_c)$ is trained by an easy-to-hard refinement loss $\mathcal{L}_{sp}$ to alleviate wrong pseudo-labels. 
% Following the easy-to-hard learning strategy, the refinement network is optimized to update noisy pseudo-labels under the supervision of $\mathcal{L_{sp}}$.
A confidence check step is introduced to construct new target set $D^*_T$ with $N^*_t$ reliable samples that satisfy the condition $\varphi(x^t,\tilde{y}^t) \geq e^{-\lambda}$. 
% filter out pseudo-labels with low confidence by comparing with a confidence threshold $P_m$. Target samples that satisfy the condition, i.e. $p^t \geq P_m$, are selected to construct pseudo-labeled train set. 
4) We do class-aware sampling and then enforce two novel losses, i.e. $\mathcal{L}_{C2C}$ and $\mathcal{L}_{P2P}$, in feature and label space to jointly align class-conditioned distribution. Step 2 to 4 are alternated in every iteration during optimization.}

\label{fig:frame}
\end{figure*}
%-------------------------------------------------------------------------

% \vspace{-3mm}
%%%%%%%%%%%%%%%%%%%%%
% {\bf Local alignment domain adaptation.}
In the more recent works, pseudo-labels are used as surrogates for the missing labels in the target domain. %pseudo-labels are exploited as estimations of the label distribution in the target domain. 
Several methods~\cite{27CondDist1,24TPN,52liang2019distant,28CondDist2,49progressive} formulate class-conditioned domain alignment by minimizing the distribution discrepancy of the prototypes.
TPN~\cite{24TPN} assigns pseudo-labels to the target samples by their nearest source prototypes and then aligns prototypical distributions.
CAN~\cite{8CAN} proposes a novel metric to model the class-level distribution discrepancy. However, these methods directly use pseudo-labels as supervision to optimize network parameters. Wrong labels can hinder the class-aware domain alignment performance. 
% In contrast, we learn from noisy labels to improve the quality of pseudo-labels.
To alleviate the wrong label problem, the semantic transfer network~\cite{27CondDist1} employs an alignment loss on centroids. ~\cite{49progressive} integrates the network with an easy-to-hard strategy to progressively select easy samples for domain alignment. Some works~\cite{56saito2017asymmetric,63zou2018unsupervised} adopt self-training by iteratively predicting pseudo-labels from the network and re-training the network with these labels. 
However, these methods generate pseudo-labels by the model trained on the source data. Consequently, the source domain bias that remains can deteriorate the quality of the pseudo-labels.
% It inevitably introduces domain shift in the network predictions, which deteriorates the reliability of pseudo-labels.
In our work, we additionally introduce an optimal assignment step to reduce noises in the pseudo-labels and a pseudo-label refinement step to improve the quality of the pseudo-labels and their confidence.
% in each iteration by training an auxiliary target-specific network. 
The experimental results empirically show the effectiveness of our approach. %in improving the class-aware domain alignment. 
% optimized over the entire target data with the easy-to-hard strategy.
% Moreover, different from previous work, our pseudo-label selection is performed by the optimized target-specific network.
% In contrast to most previous methods~\cite{49progressive,27CondDist1,28CondDist2}, besides aligning feature distributions, we also address label distribution shifts. Our proposals are empirically confirmed to achieve the state-of-the-arts on all four classification benchmarks.
% And similar to~\cite{54jiang2020implicit}, we also do class-wise sampling over labeled source data and pseudo-labeled target samples for class-level domain alignment. The intra-domain class imbalance and inter-domain class-distribution shift can thus be addressed under the uniform label distribution assumption. 

% Whereas, the difference is that our target set for sampling are composed by easy samples of high confidence. 
% Therefore a confidence check strategy is further introduced and incorporated into the pseudo-label refinement module.

%-------------------------------------------------------------------------
\section{Problem Formulation}
%\label{Method:problem}
UDA aims to deal with the domain shift problem between labeled source data $D_S=\{(x_i^s,y_i^s)\}_{i=1}^{N_s}$ and unlabeled target data $D_T=\{x_i^t\}_{i=1}^{N_t}$. Given $N_s$ source samples $D_S$ and $N_t$ target samples $D_T$ with the shared label space $\mathcal{Y}$, we expect to make prediction $\{\hat{y}^t\}$ of $\{x^t\}$ by utilizing the source label information $\{y^s\}$, where $|\mathcal{Y}|=K$ and $y^s \in \{0, 1, \dots, K-1\}$ represent the label space with $K$ classes. Based on the assumption from~\cite{10JAN,21DC}, 
there exists a shared feature space across domains, and 
the goal is to learn an domain-invariant feature embedding function $\phi(x; \theta_f): x \mapsto f$ that maps $x$ to the shared feature space and a classifier $g(f; \theta_c): f \mapsto \mathcal{Y}$ that maps features to the label space. $\{\theta_f, \theta_c\}$ are the learnable parameters of the network, denoted as a composition function $g(\phi(x; \theta_f); \theta_c)$.
% We train a shared feature extraction network and an identical classifier for both source and target data. 

%-------------------------------------------------------------------------
\vspace{-2mm}
\section{Our Method}
% We formulate the problem in Section \ref{Method:problem}. Section \ref{Method:framework} explains our framework with details of each component. Section \ref{Method:optimization} discusses the training procedure and optimization objective of our model. 
% Finally, we give some theoretical insight on the proposed loss function in Section \ref{Method:theore_analy}.
Figure~\ref{fig:frame} shows our CA-UDA framework with four main steps: 1) initialization, 2) optimal assignment, 3) pseudo-label refinement, and 4) class-aware domain alignment.

%------------------------------------------------------------------------
% \vspace{-2mm}
\subsection{Initialization}
We initialize our classification network by training with the labeled source domain data $D_S$. 
% $\phi: x \mapsto f$ is the feature embedding extractor and $g: f \mapsto \{L, p\}$ is the classification layer that maps the feature $f$ into a class label $y$ with probability $p$. $\{\theta_f, \theta_c\}$ are the learnable network parameters. 
After training, a set of feature embeddings $f^s = \{f_1^s, \dots, f_{N_s}^s\}$ for the labeled source domain data $D_S$ and $f^t = \{f_1^t, \dots, f_{N_t}^t\}$ with the corresponding pseudo-labels $\tilde{y}^t = \{\tilde{y}_1^t, \dots, \tilde{y}_{N_t}^t\}$ for the unlabeled target domain data $D_T$ are generated, where features with the same class in the same domain are potentially clustered as observed in DeepCluster~\cite{64caron2018deepcluster}.
% pseudo-labels $\tilde{y}^t = \{\tilde{y}_1^t, \dots, \tilde{y}_{N_t}^t\}$ for the unlabeled target domain data $D_T=\{x_i^t\}_{i=1}^{N_t}$. A set of feature embeddings.
% Our work is based on the deep clustering algorithms~\cite{64caron2018deepcluster} that discover clusters in training data and learn feature representation simultaneously. Each cluster in $f^t$ is assumed to be distinguishable from others and corresponds to one cluster in $f^s$ with a specific class label.
Note that we only initialize the network once.
% and aim to improve the feature discriminability for clustering. 
In the subsequent training epochs, the target domain feature embeddings $\tilde{f}^t$ and pseudo-labels $\tilde{y}^t$, and the source domain feature embeddings $f^s$ are directly obtained from the current network. 

\vspace{-1mm}
\subsection{Optimal Assignment}
The goal of the optimal assignment step is to assign pseudo-labels to the target domain feature embeddings $f^t$ using $f^s$ and its corresponding labels $y^s$. Due to the lack of strong labels for the target domain data, we apply the unsupervised k-means algorithm~\cite{Jin2010} on $f^t$ to get $K$ clusters $C^t = \{c^t_1, \dots, c^t_K\}$, where $c^t_k$ is the centroid of the $k^{\text{th}}$ cluster. Intuitively, feature embeddings that belong to the same class should get assigned to the same cluster. Nonetheless, the actual label of each cluster remains unknown at this stage since unsupervised clustering algorithms
such as k-means are based on nearest-neighbor.
% and thus is unable to determine the labels. 

The optimal assignment is used to circumvent the problem of missing labels in the target feature embedding clusters $C^t$. To this end, we first compute the centroids of the feature embeddings $C^s = \{c^s_1, \dots, c^s_K\}$ in the source domain using the known labels $y^s$, where $c^s_k$ is the centroid of the feature embeddings corresponding to the class label $k$. Subsequently, we align two sets of clusters $C^s$ and $C^t$ using optimal assignment. Formally, the optimal alignment of the two sets of clusters is formulated as a linear program: 
\begin{equation}
\begin{array}{ll@{}ll}
\text{minimize}  &\displaystyle\sum\limits_{i \in |C^s|, j \in |C^t|} d_{ij}m_{ij} &\\
\text{subject to}& \displaystyle\sum\limits_{i \in |C^t|} m_{ij} = 1, &  j \in |C^s|\\ & \displaystyle\sum\limits_{j \in |C^s|} m_{ij} = 1, &  i \in |C^t| \\
& m_{ij} \in \{0,1\}, &  i \in |C^s|, j \in |C^t|,
\end{array} \label{eq:optimalAssignment}
\end{equation}
where $d_{ij}$ is the Euclidean distance between the $i^\text{th}$ and $j^\text{th}$ centroids in the source $c_i^s$ and target $c_j^t$ domains. $m_{ij} \in \mathbf{M}$ is an element in the $K \times K$ permutation matrix $\mathbf{M}$. $m_{ij} = 1$ indicates that the cluster $c_i^s$ from the source domain and $c_j^t$ from the target domain are assigned as a match. The two summation constraints ensure the uniqueness of the cluster assignment. The objective of the optimal assignment is to find the optimal permutation matrix $\mathbf{M}$ such that the total distance between all the corresponding cluster pairs are minimized. We use the Hungarian algorithm~\cite{Kuhn55thehungarian} as solver for the optimal assignment. After the optimal alignment of the clusters, each target cluster in $C^t$ is pseudo-labeled by the label of its corresponding source cluster and we have pseudo-label set $\tilde{y}^{t} = \{\tilde{y}^{t}_{1}, \dots, \tilde{y}^{t}_{N_t}\}$.
% the label of each source domain cluster is assigned to its corresponding target domain cluster as pseudo-labels in the feature space $\tilde{y}^{t}_f = \{\tilde{y}^{t}_{f,1}, \dots, \tilde{y}^{t}_{f,N_t}\}$, which we use in the pseudo-label refinement step. 

\vspace{-4mm}
\paragraph{Remarks:} It should be noted that pseudo-labels $\tilde{y}^{t}$ in the target domain are far from ideal at this stage. In particular, these pseudo-labels are noisy and corrupted with outliers, and thus there exists a lot of samples in the target domain with discrepancies between $\tilde{y}^{t}$ and true $y^{t}$. These discrepancies are plentiful especially in the decision boundaries due to two reasons: 1) k-means is based on nearest neighbors and does not discriminate against classes; and 2) the classification network trained on the source domain does not work well on the target domain due to domain shift. 
%Nonetheless, we use the pseudo-labels obtained from the feature space $\tilde{y}^{t}_f$ in the next step since the label assignments are based on optimal alignment to the features in the source domain.  

\subsection{Pseudo-Label Refinement}
To mitigate the detrimental effects of wrong pseudo-labels, we propose an easy-to-hard pseudo-label refinement process to improve the quality of the pseudo-labels inspired by self-paced learning~\cite{39self-paced}. An auxiliary classification network $g_t(\phi_t(x; \theta^t_f); \theta^t_c)$ is trained in the target domain independently, which aims to simultaneously select easy samples and learn network parameters $\theta^t$, i.e. $\theta^t_f$ and $\theta^t_c$. According to the definition in~\cite{39self-paced}, easy samples lie far from the decision boundaries and thus their correct labels can be predicted easily. Specifically, we use the following loss function $\mathcal{L}_{sp}$ to update the network at the $n^\text{th}$ epoch:
%We first get the easy samples as the target domain data with consistent feature and output space pseudo-labels, i.e. $x^t_\text{co} = \{\dots, x^t_{\text{co},j} ,\dots \}$ such that $\mathbb{I}(\tilde{y}^{t}_{f,j} = \tilde{y}^{t}_{j}) = 1$. These are the samples that are more likely to have the correct pseudo-labels, and thus they used throughout the whole pseudo-label refinement process. 
%The classification network $g_t(\phi_t(x; \theta^t_f); \theta^t_c)$ is first trained with the target data with consistent pseudo-labels $x^t_\text{co}$. We then use the remaining inconsistent samples $x^t_\text{in}$ in a self-paced learning~\cite{} training process: 
\begin{equation}
    \begin{split}
    & (\theta^t_{n+1},v_{n+1}) =  \argmin_{\theta^t, v} \mathcal{L}_{sp}(\theta^t_{n},v_{n}), \quad \text{where}\\
    & \mathcal{L}_{sp}(\theta^t_{n},v_{n})= \frac{\mathcal{C}^1_n (\theta^t_n,v_n) + \mathcal{C}^2_n(v_n)}{|x^t|}, \\
    & \mathcal{C}^1_n = \sum_{i=1}^{|x^t|}v^i_n \mathbb{L}(x_{i}^t, \tilde{y}^t_i;\theta^t_n),\quad
    \mathcal{C}^2_n = -\gamma^{n}\lambda \sum_{i=1}^{|x^t|}v^i_n. 
    \end{split} 
\label{eq:LossSelfPaced}
\end{equation}
$\mathbb{L}(\cdot)$ is the negative log-likelihood as in~\cite{53guo2017calibration} to measure confidence, i.e. the probability of correctness, and $v_n^i \in\{0,1\}^{|x^t|}$ is a binary variable to indicate whether the sample $x_{i}$ is included for minimization. $\gamma^{n}\lambda$ is the confidence threshold at the $n_{th}$ epoch to control the target samples to be selected. We initially set the threshold with a low value (i.e. $\lambda$) and gradually increase it by multiplying with a growing factor $\gamma$. $n=\{0,1,2,...,N_{max}\}$ is the training epoch and $\gamma$ is a constant. It can be seen that samples with high likelihood $\varphi=e^{-\mathbb{L}(\cdot)}$ are considered the easy samples and selected with the condition $\varphi \geq e^{-\gamma^n \lambda}$. The selection gradually expands to the more difficult samples as $n$ increases. The parameter update proceeds until the entire target data are used and the maximum epoch is reached, i.e. $v^i_n=1,\forall i$ and $n= N_{max}$. The optimum parameters $\theta^{t*}$ are cached for continuous update in the next iteration. To alleviate the wrong pseudo-label problem for domain alignment, we perform a confidence-check step to only select easy samples in $D_T$, which satisfy the condition $\varphi(x_i^t,\tilde{y}_i^t;\theta^{t*}) \geq e^{-\lambda}$, to construct new target training set $D_T^*=\{(x_i^t,\tilde{y}_i^t)\}_{i=1}^{N_t^*}$ with $N_t^*$ samples. Specifically, we set $n=0$ for simplicity and use $\lambda$ as the confidence threshold to filter out hard samples which lie near the decision margins in the target domain. 
% We update the pseudo-labels for all the target domain samples with the classification network $g_t(\phi_t(x; \theta^t_f); \theta^t_c)$ trained with the easy-to-hard process.

\vspace{-4mm}
\paragraph{Remarks:}  In curriculum~\cite{41curriculum} and self-paced learning~\cite{39self-paced}, it is observed that a deep network overfits quickly to samples with bad labels in training data. It is thus beneficial to first train the network with easier samples, i.e. samples with potentially correct labels, before adding the harder samples. As discussed in the previous section, noisy pseudo-labels still exist from the optimal assignment step. %As discussed in the previous section, the wrong pseudo-labels arise from: 1) k-means for intra-domain unlabeled target data, and 2) the domain gap between the corresponding source and target prototypes.  
% it is necessary to utilize the easy-to-hard training strategy on instances $x^t$ without domain shift to update parameters $(\theta^t_f), \theta^t_c)$, which can improve the quality of final intra-domain pseudo-labels. 
% Considering the trade-off that dense pseudo-labels tend to be much more noisy while sparse labels by cutting-out a subset of unconfident labels are prone to miss meaningful information, we thus refine the entire pseudo-label set first and then do pseudo-label selection for class-aware domain alignment in each iteration.
It is therefore necessary to independently train a target-specific network on instances $x^t$ aiming to decouple domain bias in the obtained pseudo-label predictions. The quality of the entire pseudo-labels can be improved with the easy-to-hard learning scheme.
% To further alleviate the wrong pseudo-labels in our class-aware domain alignment, the confidence check performs pseudo-label selection by favoring easy samples with high confidence. 
Furthermore, the confidence check helps to eliminate errors from the wrong labels by favoring easy samples with high confidence.
% ones that lie far from the margins (easy samples with high likelihood) can be rectified by the class-conditioned sampling on $D^*_T$. 
% As a result, the training of this network improves the decision boundaries specifically in the target domain, and therefore improving quality of the final pseudo-labels. %Interestingly, our easy-to-hard pseudo-label refinement step can be also be seen as self-supervised learning.  

\subsection{Class-Aware Domain Alignment}
\paragraph{Cross-domain inter-clusters.} Once we get the refined pseudo-labels from the previous step, we do a class-aware domain alignment to close the domain gap while preserving the class information. Under the assumption that the shared label space follows uniform distribution as in~\cite{54jiang2020implicit}, 
% a subset class labels in current training batch $\mathcal{Y_B}\subset \mathcal{Y}$ are firstly sampled and mini-batch 
samples conditioned on labels are then drawn in both source and target domains ($D_S$ and $D^{*}_T$).
To this end, we propose a modified MMD loss on the feature embeddings that considers the multiple means formed by the different clusters of classes. We call this the center-to-center loss, i.e. 
\begin{align}
\begin{autobreak}
\MoveEqLeft
\mathcal{L}_{C2C} = \frac{1}{K}\sum_{k=1}^{K}  \Bigg\| \frac{1}{N_s^k} \sum_{i=1}^{N_s^k}\phi(x_i^{s,k}) - \frac{1}{N_t^k} \sum_{j=1}^{N_t^k}\phi(x_j^{t,k}) \Bigg\|^2_{\mathcal{H}}
~~= \frac{1}{K}\sum_{k=1}^K\{\frac{1}{{N_s^k}^2}\sum_{i=1}^{N_s^k}\sum_{j=1}^{N_s^k}\mathcal{K}(\phi(x_i^{s,k}),\phi(x_j^{s,k}))
+\frac{1}{{N_t^k}^2}\sum_{i=1}^{N_t^k}\sum_{j=1}^{N_t^k}\mathcal{K}(\phi(x_i^{t,k}),\phi(x_j^{t,k}))
-\frac{2}{N_s^k N_t^k}\sum_{i=1}^{N_s^k}\sum_{j=1}^{N_t^k}\mathcal{K}(\phi(x_i^{s,k}),\phi(x_j^{t,k}))\},
\end{autobreak} \label{eq:CA-UDALoss}
\end{align}
where $\phi(x_i^{s,k})\in D_S$ and $\phi(x_j^{t,k})\in D^{*}_T$ are the source and target feature embeddings (omitting $\theta_f$ from $\phi(x;\theta_f)$) with the $k^\text{th}$ class label, respectively. $\mathcal{K}(.,.)$ is the radial basis kernel function (RBF) and $\|.\|_{\mathcal{H}}$ is the reproducing kernel Hilbert space. The inner terms in the first line of Eq.~\ref{eq:CA-UDALoss} denote the square-difference between the two empirical means from the $k^\text{th}$ class of the source and target domains. Putting the inner terms into the summation over $K$ classes gives us the square-difference between the empirical means of all corresponding classes in the source and target domains. The minimization of $\mathcal{L}_{C2C}$ is equivalent to the minimization of the inter-cluster center-to-center distances between the source and target domains. In our experiments, two batches of samples in an epoch: $B_S=\{(x_i^{s,k},y_i=k)\}_{i=1}^{N_s^k}$ and $B_T=\{(x_i^{t,k},\tilde{y}_i=k)\}_{i=1}^{N_s^k}$ are sampled from $D_S$ and $D^{*}_T$ according to the class label $k \in \mathcal{Y_B}\subset \mathcal{Y}$. The mini-batch label space $\mathcal{Y_B}$ consists of $K_B$ classes, i.e. $|\mathcal{Y_B}|=K_B$. Each class has ${N_s^k}$ source and ${N_t^k}$ target samples, respectively. The labels in $\mathcal{Y_B}$ to be aligned are uniformly chosen from the shared label space $\mathcal{Y}$ to ensure a shared distribution in the corresponding empirical means across domains. Consequently, the class imbalance within the domain and class-distribution shift across domains can be mitigated from the sampling perspective.

\vspace{-4mm}
\paragraph{In-domain intra-clusters.} In similar vein, we introduce another probability-to-probability loss on the output probability of the classification layer, i.e. 
%\yipin{(a bit confused why this p2p loss could reduce in-domain intra-clusters distance? Feel like it's doing similar things as C2C loss.) \textbf{Ans: CA-UDA pulls the two multimodal distributions together, i.e. close the domain gap. p2p does classification, i.e. make the decision boundaries clearer. See the last parts of previous and this paragraphs.}}  
\begin{equation}
    \mathcal{L}_{P2P} = \frac{1}{K}\sum_{k=1}^{K}  \Bigg\| \frac{1}{N_s^k} \sum_{i=1}^{N_s^k}p_i^{s,k} - \frac{1}{N_t^k} \sum_{j=1}^{N_t^k}p_j^{t,k} \Bigg\|^2_{\mathcal{H}}, \label{eq:P2PLoss}
\end{equation}
where $p_i^{s,k}$ and $p_j^{t,k}$ are the probabilities of a source $x_i^s$ and target $x_j^t$ domain samples taking the $k^\text{th}$ class label, respectively. 
% Minimizing $\mathcal{L}_{P2P}$ has the effect of forcing the source and target domain samples in the same class to take the same probability values. 
$\mathcal{L}_{P2P}$ aims to minimize the probability difference of the source and target samples in the same class.
It is important that $\mathcal{L}_{P2P}$ has to be minimized together with the cross-entropy loss $\mathcal{L}_\text{CE}(x^s, y^s)$ from the fully supervised source domain. Specifically, $\mathcal{L}_\text{CE}(x^s, y^s)$ provides a constraint on the probability values in Eq.~\ref{eq:P2PLoss} that forces the in-domain feature embeddings into more discriminative clusters. This implies that in addition to maximizing the in-domain inter-cluster distances of the feature embeddings, the in-domain intra-cluster distances are concurrently minimized towards shaping a better decision boundary. 

%-------------------------------------------------------------------------
\subsection{Optimization.}
\label{Method:optimization}
%The framework of our CA-UDA method is illstrated in Fig. \ref{fig:frame}. This section mainly discuss the optimization procedure in three steps by alternating.

The optimization of our classification network for UDA is performed by an initialization step and an alternating three-stage training procedure over last three steps. The first initialization step is only performed once.

\vspace{-3mm}
\paragraph{1) Initialization.} We initialize our network by doing a fully supervised pre-training on the labeled source data, where we minimize the cross-entropy loss over the network parameters $\theta = \{\theta_f, \theta_c\}$:
\begin{equation}
    \mathcal{L}_\text{CE} = -\E_{(x^s,y^s)\thicksim D_S} \sum_{k=1}^K \mathbbm{1}_{c=y^s}log(y|x^s).
\end{equation}

% Firstly, we pretrain our network with labeled source data only to obtain task-specific discriminative feature representations. The objective of this step is to minimize the cross entropy classification loss with the network parameterized by $\theta$, which is as follows:
% \begin{equation}
%     \min_{\theta}L_{ce} = -\E_{(x_s,y_s)\thicksim(D_S, Y_S)} \sum_{c=1}^C \mathbbm{1}_{c=y_s}log(y|x_s)
% \end{equation}

\vspace{-3mm}
\paragraph{2) Optimal assignment.} 
% The k-means, optimal assignment (cf. Eq.~\ref{eq:optimalAssignment}) and easy-to-hard pseudo-label refinement loss (cf. Eq.~\ref{eq:LossSelfPaced}) in the optimal assignment and pseudo-label refinement steps are done at the beginning of every epoch during training.
At the start of every training iteration, the k-means and optimal assignment are performed, where
% We use a modified k-means algorithm to do clustering in the target domain. 
the initial centroids of the k-means on the unlabeled target domain data are generated from the the labeled source cluster centroids. Subsequently, we cache the centroids from the previous epoch into the memory and update them by a simple moving average in the current epoch:
\begin{equation}
c_k \leftarrow \frac{\phi_k(x; \theta_f)}{\| \phi_k(x; \theta_f) \|}_2 + \alpha \cdot c_k,
\vspace{-3mm}
\end{equation}
to ensure stability and efficiency. $\alpha$ is the update momentum coefficient set to 1 for simplicity. $\phi_k(x; \theta_f)$ is the feature embedding assigned to the $k^\text{th}$ cluster from the nearest neighbor search. 
% As mentioned in the previous section, 
We use the Hungarian algorithm~\cite{Kuhn55thehungarian} as the solver for the optimal assignment problem formulated as a linear program in Eq.~\ref{eq:optimalAssignment}. 
% For the pseudo-label refinement task, we train an auxiliary classification network $g_t(\phi_t(x; \theta^t_f); \theta^t_c)$ for the unlabeled target domain using the easy-to-hard loss $\mathcal{L}_{sp}$ from Eq.~\ref{eq:LossSelfPaced}.
% This auxiliary network is discarded after training. 

\vspace{-3mm}
\paragraph{3) Pseudo-label refinement.}
% At the first epoch of every iteration,
Parameters $\theta^t_f$ and $\theta^t_c$ of the auxiliary network $g_t(\phi_t(x; \theta^t_f); \theta^t_c)$ are updated based on the pseudo-labels obtained from the previous optimal assignment step. The easy-to-hard refinement loss $\mathcal{L}_{sp}$ from Eq.~\ref{eq:LossSelfPaced} is iteratively minimized by gradually increasing the threshold $\gamma^n\lambda$ with the iterations, until all target samples are included and the maximum epoch is reached. The optimum parameters $\theta^{t*}$ are cached for continuous optimization in the next iteration. 
Finally, the confidence check selects reliable examples 
that pass the confidence threshold $\lambda$ 
to construct a new training target set $D_T^*=\{(x_i^t,\tilde{y}_i^t)\}_{i=1}^{N_t^*}$, i.e. $e^{-\mathbb{L}(x_i^t,\tilde{y}_i^t;\theta^{t*})} \geq e^{-\lambda}$.
% is adopted to rectify falsely pseudo-labeled samples near the decision boundaries by selecting easy examples with high confidence to construct new training target set for domain alignment. 
% after the refinement step to filter out hard target samples. A confidence threshold $P_m$ (in Fig. \ref{fig:frame}) is compared with confidence values of all pseudo-labeled target samples, and only those with high confidence (i.e., $p^t > P_m$) are used for the next step class-aware domain alignment.

\vspace{-3mm}
\paragraph{4) Class-aware domain alignment.} 
% The class-aware domain alignment is done at every iteration during training. 
For training efficiency, we uniformly sample a subset classes from which $n$ samples are randomly drawn in the source and target domain, i.e. $x^k = \{x_1^k, ...x_n^k\},~\forall K$, where $x_n^k \sim (D_S^k, D_T^k)$.
The total loss $\mathcal{L}_\text{DA}$ we used to train the classification network $g(\phi(x;\theta_f); \theta_c)$ consists of the center-to-center $\mathcal{L}_\text{C2C}$ (cf. Eq.~\ref{eq:CA-UDALoss}), probability-to-probability $\mathcal{L}_\text{P2P}$ (cf. Eq.~\ref{eq:P2PLoss}), and source domain cross-entropy $\mathcal{L}_\text{CE}(x^s,y^s)$ loss terms:
\begin{equation}
    \mathcal{L}_\text{DA} = \mathop\mathbbm{E}_{x^1, \dots, x^K} \{ \tau_1\mathcal{L}_\text{C2C} + \tau_2\mathcal{L}_\text{P2P} + \mathcal{L}_\text{CE}(x^s,y^s) \}, \label{eq:DALoss}
\end{equation}
where $\tau_1$ and $\tau_2$ are hyperparameters to balance the losses.

%------------------------------------------------------------------------
\section{Experiments}
We conduct experiments on four standard benchmark datasets: 1) Office-31~\cite{3dataoffice}, 2) ImageCLEF-DA~\cite{5dataclef}, 3) VisDA-2017~\cite{4datavisda} and 4) Digit-Five to evaluate the performance of our proposed CA-UDA framework in comparison to other existing state-of-the-art methods. In addition, we show ablation studies to evaluate the effectiveness of each component in our framework.

\subsection{Datasets and Experimental Setting}

\paragraph{Datasets.}
\textbf{Office-31}~\cite{3dataoffice} includes three domains: Amazon (A), DSLR (D) and Webcam (W), and contains a total of 4,110 images covering 31 categories. A combination of six pairs of source-target domain settings are evaluated. \textbf{ImageCLEF-DA}~\cite{5dataclef} includes three domains: Caltech-256 (C), ImageNet ILSVRC 2012 (I), and Pascal VOC 2012 (P) with 12 categories, where each category contains equal number of 50 images. \textbf{VisDA-2017} \cite{4datavisda} is a challenging dataset 
due to the big domain shift between the synthetic images (152,397 images from VisDA) and the real images (55,388 images from COCO). We evaluate our method on the setting of synthetic-to-real as the source-to-target domain. To make a fair comparison with other methods, We choose the following three digit datasets: MNIST (M)~\cite{Digit-M}, USPS (U)~\cite{Digit-U} and SVHN (S)~\cite{Digit-S} in \textbf{Digit-Five} as different domains. MNIST (M) and USPS (U) are handwritten Digits containing 70K and 9.3K images, respectively. SVHN has 100K colored digits collected from the real world.

\vspace{-3mm}
\paragraph{Implementation details.}
For object classification datasets, we adopt ResNet-50~\cite{1resnet50} and Resnet-101 \cite{2resnet101} pretrained on the ImageNet as the backbone. The last FC layer is replaced with one FC layer of task-specific dimensions. The network parameters are all shared between source and target domain except for the batch normalization layers. We fine-tune the network feature extractor and train the classifier from scratch. For digit classification, we follow the network architecture in~\cite{12RevGrad} and~\cite{7lenet2}, where there are only two convolutional layers as the feature extractor. We use stochastic gradient descent (SGD) with momentum of 0.9 and weight decay of 0.0005 to train the network. 
We set the learning rate schedule $\eta_p = \frac{\eta_0}{(1+\alpha t)^\beta}$ as in~\cite{12RevGrad,14DANN,8CAN,10JAN,13MADA,23ADDA}. Specifically, the iteration $t$ linearly changing from 0 to 1, $\alpha =10$ \& $\beta = 0.75$. Following~\cite{8CAN,23ADDA}, $\beta = 2.25$ for VisDA-2017 dataset, $\eta_0=0.001$ for feature extractor, $\eta_0=0.01$ for task-specific classifier.
We empirically set
the hyperparameters in Eq.~\ref{eq:LossSelfPaced} to
$\lambda=0.1, \gamma=1.3$. 
This corresponds to the initial likelihood $e^{-\lambda}\simeq 0.9$ that decreases with an exponential decay $(\cdot)^{\gamma}$. 
For $\tau_1$ and $\tau_2$ in Eq.~\ref{eq:DALoss}, a sensitively analysis is done on VisDA-2017, and
the best performing $\tau_1=\tau_2=0.3$ is obtained. 
% We show above sensitive analysis in experiments.

%------------------------------------------------------------------------
\vspace{-3mm}
\paragraph{Baselines.} We compare our CA-UDA with several baselines to verify its effectiveness: 1) \textbf{Source-only}.
%  optimizes the network with labeled data in the source domain only. 
2) Domain-level alignment methods. Representative baselines include: discrepancy-based models, i.e. \textbf{DAN}~\cite{9DAN} and \textbf{JAN}~\cite{10JAN},
% which minimize distribution discrepancy across domains. 
and adversarial-discriminative models, i.e.
\textbf{DANN}~\cite{14DANN}, \textbf{MADA}~\cite{13MADA}, \textbf{MCD}~\cite{16MCD} and \textbf{RevGrad}~\cite{12RevGrad}.
% , which adopt adversarial training to learn domain-invariant features. 
3) Class conditional alignment methods which utilize pseudo-labels as supervision in the target domain, i.e. \textbf{TPN}~\cite{24TPN},
% assigns each target sample a pseudo-label with the nearest prototype in the source domain.
\textbf{CAN}~\cite{8CAN} and \textbf{A$^2$LP}~\cite{20A2LP}, and several baselines that also consider to solve the
% improve the quality of pseudo-labels and suppress noises conveyed by the wrong pseudo-labeled samples
wrong pseudo-label problem, i.e.~\textbf{CAT}~\cite{28CondDist2},~\textbf{MSTN}~\cite{27CondDist1} and~\textbf{PFAN}~\cite{49progressive}.
% use source prototypes to label the target data and adopt a prototype alignment loss to minimize the conditional distribution discrepancy. 
% \textbf{CAT}~\cite{28CondDist2} builds a teacher model to provide pseudo-labels and proposes a cluster alignment loss to align class-conditional distributions.
% \textbf{CAN}~\cite{8CAN} explicitly models the intra-class and inter-class domain discrepancy for optimization. \textbf{A$^2$LP}~\cite{20A2LP} is a variant of Label Propagation (LP)~\cite{11LP} that tackles domain shift problem by generating virtual unlabeled instances with high-confidence label estimation to improve the quality of pseudo-labels. 

%------------------------------------------------------------------------

\subsection{Comparison with Baselines}
%------------------------------------------------------------------------
\paragraph{Digit classification.}
We evaluate our CA-UDA framework on three source-to-target domain adaptation settings from the Digit-Five dataset: M $\to$ U, U $\to$ M and S $\to$ M.
Table~\ref{tab: digit} shows that
% the classification results of all baseline methods (taken from~\cite{24TPN}) on these digit datasets. 
our method achieves state-of-the-art performance in all three domain adaptation subtasks. Compared with DAN~\cite{9DAN} and JAN~\cite{10JAN}, 
% which also explicitly measure discrepancy across domains,
class-level alignment methods generally perform much better. 
% than the domain-level alignment methods, e.g. JAN~\cite{10JAN} and \textbf{DANN}~\cite{14DANN}.
It indicates that matching distribution for each category helps to improve domain alignment. In comparison to TPN that uses prototype alignment, 
% and CA-UDA follow similar pipeline that uses clustering for pseudo-label generation and a class-level prototype alignment loss, 
our CA-UDA is observed to achieve higher accuracy, e.g. in S $\to$ M, our CA-UDA achieves +4.63\% and + 5.49\% with $\mathcal{L}_{sp}$. It implies the effectiveness of the optimal assignment and pseudo-label refinement in our CA-UDA in reducing the negative effects from wrong pseudo-labels. Furthermore, our model outperforms or shows comparable performance with the pseudo-labeling baselines. This confirms the capability of our proposals for mitigating domain shift in pseudo-label predictions.
% Therefore,
% two novel class-aware alignment losses in both feature and label space. 
% the superiority of our model is mostly contributed by alleviating wrong pseudo-labels in the target domain.
% , especially with an auxiliary pseudo-label refinement loss.
   
% two advantages of our model: 1) the proposed class-aware C2C and P2P losses based on the class-sampling technique are more effective in dealing with misclassification problem in the target domain. 2) The pseudo-label refinement strategy is essential in conditional alignment-based DA methods since the quality of target pseudo labels is closely related to the performance of class-level alignment. Furthermore, our method achieves state-of-the-art performance in all three domain adaptation subtasks.

\begin{table}
\begin{center}
\scalebox{0.8}{
\begin{tabular}{c|cccc}
\hline
Method & M $\to$ U & U$\to$ M & S $\to$ M & mean \\
\hline
Source-only & 75.2 & 57.1 & 60.1 & 64.1\\
RevGrad~\cite{12RevGrad} & 77.1& 73.0 & 73.9 & 74.7\\
DAN~\cite{9DAN} & 80.3 & 77.8 & 73.5 & 77.2\\
% RTN \cite{22RTN} & 82.0 & 81.2 & 75.3 \\
JAN~\cite{10JAN}& 84.4 & 83.4 & 78.4 & 82.1\\
MCD~\cite{16MCD}& 90.0 & 88.5 & 83.3 & 87.3\\
DANN~\cite{14DANN} & 90.4 & 94.7 & 84.2 & 89.8\\
TPN~\cite{24TPN}& 92.1 & 94.1 & 93.0 & 93.1\\
MSTN~\cite{27CondDist1} & 92.9 & - & 91.7 & - \\
PFAN~\cite{49progressive} & 95.0 & - & 93.9 & - \\
rRevGrad+CAT~\cite{28CondDist2} & 94.0 & 96.0 & \textbf{98.8} & 96.27\\
\hline
Ours (without $\mathcal{L}_{sp}$) & 93.79 & 96.62 & 97.63 & 96.01\\
Ours (with $\mathcal{L}_{sp}$) & \textbf{95.43} & \textbf{97.22} & 98.49 & \textbf{97.05}\\
\hline
\end{tabular}}
\end{center}
\caption{Classification accuracy (\%) on Digit-five. %Our methods named "without $\mathcal{L}_{sp}$" and "with $\mathcal{L}_{sp}$" refer to the training with C2C loss and additionally pseudo-label refinement classifier, respectively
}
\label{tab: digit}
\end{table}
%------------------------------------------------------------------------

%------------------------------------------------------------------------
\begin{table}
\begin{center}
\scalebox{0.6}{
\begin{tabular}{c|ccccccc}
\hline
Method &A $\to$ W & W $\to$ A & A $\to$ D & D $\to$ A & W $\to$ D & D $\to$ W & mean\\
\hline
Source-only & 68.4 & 60.7 & 68.9 & 62.5 & 99.3 & 96.7 & 76.1\\
RevGrad~\cite{12RevGrad} & 82.0 & 67.4 & 79.7 & 68.2& 99.1 & 96.9 & 82.2\\
DAN~\cite{9DAN} & 80.5 & 62.8 & 78.6 & 63.6 & 99.6 & 97.1 & 80.4\\
JAN~\cite{10JAN}& 85.4 & 70.0 & 84.7 & 68.6 & 99.8 & 97.4 & 84.3\\
MADA~\cite{13MADA}& 90.0 & 66.4 & 87.8 & 70.3 & 99.6 & 97.4 & 85.3\\
MSTN~\cite{27CondDist1}& 91.3 & 65.6 & 90.4 & 72.7 & 100.0 & 98.9 & 86.5\\
rRevGrad+CAT~\cite{28CondDist2}& 94.4 & 70.2 & 90.8 & 72.2 & 100.0 & 98.0 & 87.6\\
CAN~\cite{8CAN}& 94.5 & 77.0 & 95.0 & 78.0 & 99.8 & 99.1 & 90.6\\
A$^2$LP~\cite{20A2LP} & 93.4 & 77.6 & 96.1 & 78.1 & \textbf{100} & 98.8 & 90.7\\
\hline
Ours (without $\mathcal{L}_{sp}$) & 93.96 & 77.21 & 93.37 & 78.31 & \textbf{100} & 98.36 & 90.20\\
Ours (with $\mathcal{L}_{sp}$) & \textbf{96.23} & \textbf{85.05} & \textbf{98.19} & \textbf{85.34} & \textbf{100} & \textbf{99.25} & \textbf{94.01}\\
\hline
\end{tabular}
}
\end{center}
\caption{Accuracy (\%) on Office-31 (ResNet-50). %Our methods named "without $\mathcal{L}_{sp}$" and "with $\mathcal{L}_{sp}$" refer to the training with C2C loss and additionally pseudo-label refinement classifier, respectively.
}
\label{tab: office}
\end{table}

%------------------------------------------------------------------------
\begin{table}
\begin{center}
\scalebox{0.6}{
\begin{tabular}{c|ccccccc}
\hline
Method &I $\to$ P & P $\to$ I & I $\to$ C & C $\to$ I & C $\to$ P & P $\to$ C & mean\\
\hline
Source-only & 74.8 & 83.9 & 91.5 & 78 & 65.5 & 91.2 & 80.8\\
DAN~\cite{9DAN}& 74.5 & 82.2 & 92.8 & 86.3 & 69.2 & 89.8 & 82.5\\
DANN~\cite{14DANN}& 75.0 & 86.0 & 96.2 & 87.0 & 74.3 & 91.5 & 85.0\\
JAN~\cite{10JAN}& 76.8 & 88.0 & 94.7 & 89.5 & 74.2 & 91.7 & 85.8\\
rRevGrad+CAT~\cite{28CondDist2}& 77.2 & 91.0 & 95.5 & 91.3 & 75.3 & 93.6 & 87.3\\
A$^2$LP~\cite{20A2LP}& 79.8 & 94.3 & 97.7 & 93.0 & 79.9 & 96.9 & 90.3\\
\hline
Ours (without $\mathcal{L}_{sp}$) & 78.67 & 94.50 & 98.00 & 94.00 & 79.33 & 97.67 & 90.36\\
Ours (with $\mathcal{L}_{sp}$) & \textbf{83.50} & \textbf{96.83} & \textbf{98.83} & \textbf{97.17} & \textbf{82.67} & \textbf{99.00} & \textbf{93.00}\\
\hline
\end{tabular}
}
\end{center}
\caption{Accuracy (\%) on ImageCLEF-DA (ResNet-50).}
\label{tab: clef}
\end{table}

%------------------------------------------------------------------------

\vspace{-3mm}
\paragraph{Object classification.}
We further evaluate our CA-UDA on the object classification datasets. % and results are shown in Table \ref{tab: office}, Table \ref{tab: clef}, Table \ref{tab: visda}, respectively.
Table \ref{tab: office} evidently shows the improved accuracy over the six transfer directions across three domains on {\bf Office-31}. 
% Our method evidently improves the classification accuracy in all transfer directions, even in the harder scenarios with substantial discrepancy, i.e. D $\to$ A and W $\to$ A. 
Compared to A$^2$LP~\cite{20A2LP}, our CA-UDA with $\mathcal{L}_{sp}$ boosts performance by 7.45\% and 7.24\% in W $\to$ A (85.05\%) and D $\to$ A (85.34\%), respectively. These results highlight the advantage of our pseudo-label refinement step, which improves the quality of pseudo-labels in the target domain for better class-aware domain alignment. Although this dataset has class imbalance within each domain and in-class data imbalance across domains, our CA-UDA still consistently achieves the best performance. 
%which is more possibly supported by the class-level sampling. 
We conjecture that it is our class-level sampling that makes our model effective on the imbalanced data.
% strategy employed in the shared label space, from which both source and target samples are drawn. By restricting the uniform label distribution prior, 
% which effectively mitigate the class imbalance and class-distribution shift.

\begin{table}
\begin{center}
\scalebox{0.8}{
\begin{tabular}{c|cc}
\hline
Method & Acc. (ResNet50) & Acc.(ResNet101) \\
\hline
Source-only & 45.6 & 50.8 \\
DAN~\cite{9DAN}& 53.0 & 61.1 \\
DANN~\cite{14DANN}& 55.0 & 57.4 \\
MCD~\cite{16MCD}& - & 71.9 \\
% CDAN+E \cite{15CDAN}& 70.0 & - \\
% LPJT \cite{19LPJT}& - & 74.0 \\
% DADA \cite{17DADA}& - & 79.8 \\
% Lee \etal \cite{18lee}& 76.2 & 81.5 \\
CAN~\cite{8CAN}& - & 87.2 \\
A$^2$LP~\cite{20A2LP}& 86.5 & 87.6 \\
\hline
Ours (without $\mathcal{L}_{sp}$) & 85.45 & 86.44 \\
Ours (with $\mathcal{L}_{sp}$) & \textbf{87.13} & \textbf{88.72} \\
\hline
\end{tabular}
}
\end{center}
\caption{Classification accuracy (\%) on VisDA-2017.}% Our methods named "without $\mathcal{L}_{sp}$" and "with $\mathcal{L}_{sp}$" refer to the training with C2C loss and additionally pseudo-label refinement classifier, respectively.}
\label{tab: visda}
\end{table}

%------------------------------------------------------------------------

\begin{table*}
\begin{center}
\scalebox{0.85}{
\begin{tabular}{c|cccccccccccc|c}
\hline
Method & plane & bcycl & bus & car & horse & knife & mcycl & person & plant & sktbrd & train & truck & mean\\
\hline
Source-only & 70.6 & 51.8 & 55.8 & 68.9 & 77.9 & 7.6 & 93.3 & 34.5 & 81.1 & 27.9 & 88.6 & 5.6& 55.3\\
% RevGrad \cite{12RevGrad} & 75.9 & 70.5 & 65.3 & 17.3 & 72.8 & 38.6 & 58.0 & 77.2 & 72.5 & 40.4 & 70.4 & 44.7 & 58.6 \\
% DC \cite{21DC}& 63.6 & 38.4 & 71.2 & 61.4 & 71.4 & 10.9 & 86.6 & 43.5 & 70.2 & 47.7 & 79.8 & 21.6 & 55.5 \\
DAN~\cite{9DAN}& 68.1 & 15.4 & 76.5 & \textbf{87.0} & 71.1 & 48.9 & 82.3 & 51.5 & 88.7 & 33.2 & \textbf{88.9} & 42.2 & 62.8 \\
% RTN \cite{22RTN}& 79.5 & 59.6 & 78.0 & 47.4 & 82.7 & 82.0 & 84.7 & 54.7 & 81.6 & 34.5 & 74.2 & 6.6 & 63.8 \\
JAN~\cite{10JAN}& 75.7 & 18.7 & 82.3 & 86.3 & 70.2 & 56.9 & 80.5 & 53.8 & 92.5 & 32.2 & 84.5 & 54.5 & 65.7 \\
MCD~\cite{16MCD}& 87.0 & 60.9 & 83.7 & 64.0 & 88.9 & 79.6 & 84.7 & 76.9 & 88.6 & 40.3 & 83.0 & 25.8 & 71.9 \\
TPN~\cite{24TPN}& 93.7 & 85.1 & 69.2 & 81.6 & 93.5 & 61.9 & 89.3 & 81.4 & 93.5 & 81.6 & 84.5 & 49.9 & 80.4 \\
CAN~\cite{8CAN}& 97.9 & \textbf{87.2} & 82.5 & 74.3 & 97.8 & 96.2 & 90.8 & 80.7 & 96.6 & \textbf{96.3} & 87.5 & 59.9 & 87.2 \\
\hline
% Ours (without $\mathcal{L}_{sp}$) & \textbf{96.13} & \textbf{87.19} & 79.74 & 65.97 & \textbf{96.01} & 96.92 & 87.42 & 86 & 95.01 & 92.06 & 88.86 & 54.18 & 85.45 \\ ResNet-50
Ours (without $\mathcal{L}_{sp}$) & 97.48 & 86.96 & 83.6 & 76.73 & \textbf{97.48} & \textbf{97.40} & 89.73 & 79.75 & 96.97 & 94.43 & 88.60 & 48.14 & 86.44 \\
% Ours (with $\mathcal{L}_{sp}$) & 96.08 & 75.31 & 80.09 & \textbf{82.91} & 88.62 & \textbf{97.64} & \textbf{93.00} & \textbf{88.23} & \textbf{96.37} & \textbf{94.34} & \textbf{92.80} & \textbf{60.13} & \textbf{87.13} \\ ResNet-50
Ours (with $\mathcal{L}_{sp}$) & \textbf{97.92} & 84.14 & \textbf{87.25} & 81.51 & 95.97 & 97.25 & \textbf{92.7} & \textbf{87.65} & \textbf{96.99} & 94.52 & 88.29 & \textbf{60.49} & \textbf{88.72} \\
\hline
\end{tabular}
}
\end{center}
\caption{Classification accuracy (\%) on VisDA-2017 (ResNet-101).} %Our methods named "without $\mathcal{L}_{sp}$" and "with $\mathcal{L}_{sp}$" refer to the training with C2C loss and additionally pseudo-label refinement classifier, respectively.}
\label{tab: visdacat}
\end{table*}

%------------------------------------------------------------------------
Table \ref{tab: clef} reports results on six adaptation tasks between pairwise domains from {\bf ImageCLEF-DA}. Our CA-UDA with $\mathcal{L}_{sp}$ significantly outperforms all existing methods on all transfer directions.
% , even on the harder I $\to$ P and C $\to$ P. 
In contrast to other datasets, the number of samples in each category are identical in all domains. Therefore, the good results reveal the scalability of our CA-UDA to different datasets.

% In comparison to the discrepancy-based DAN~\cite{9DAN} and JAN~\cite{10JAN}, our CA-UDA shows better performance in closing the domain gap at class-level. It demonstrates the superiority of our class-aware domain alignment approach over naive global alignment. Furthermore, 
%\yipin{(Should clarify which backbone is applied for the above object classification exps, ResNet50 or ResNet101?) \textbf{Ans: Have added this in the caption of every table.}}

%------------------------------------------------------------------------

Table \ref{tab: visda} lists the classification accuracy on {\bf VisDA-2017}, where all the baseline results are directly taken from~\cite{20A2LP}. We evaluate CA-UDA with two backbone networks, i.e. ResNet-50 and ResNet-101. %Although there is still a small gap between our CA-UDA and $A^LP$, the achieved good performance indicates our method is also effective on the large dataset.
CA-UDA with $\mathcal{L}_{sp}$ outperforms all baselines. Table \ref{tab: visdacat} summarizes the accuracy over 12 categories. Prediction bias towards some classes can be observed as the performing of all baselines fluctuates over different categories. Our method achieves the best performance on most categories and the highest average accuracy.  %due to the cluster misalignment during the optimal assignment stage. This indicates the importance of the careful design to deal with the label refinement stage, which give the wrongly labeled samples chances to be corrected after the initial assignment. 
%------------------------------------------------------------------------

% \begin{figure*}[t!]
% \centering
% % \quad
% \subfigure[Source-only]{
% \begin{minipage}[b]{0.3\textwidth}
%     \includegraphics[width=5.5cm]{iccv2021_CAUDA/LaTeX/figures/tgt_conf_source_only.eps}
% \end{minipage}
% }
% \quad
% \subfigure[MMD]{
% \begin{minipage}[b]{0.3\textwidth}
%     \includegraphics[width=5.5cm]{iccv2021_CAUDA/LaTeX/figures/tgt_conf_mmd.eps}
% \end{minipage}
% }
% \quad
% \subfigure[CA-UDA with $\mathcal{L}_{sp}$]{
% \begin{minipage}[b]{0.3\textwidth}
%     \includegraphics[width=5.5cm]{iccv2021_CAUDA/LaTeX/figures/tgt_conf_C2C.eps}
% \end{minipage}
% }
% \caption{The visualization results of target features and confusion matrix on the VisDA-2017 dataset. Three models are: (a) Source-only, (b) MMD, (c) CA-UDA with $\mathcal{L}_{sp}$. Labels from 0 to 11 correspond to the order of all categories listed in Table \ref{tab: visdacat}.}
% \label{fig: feat_conf}
% \end{figure*}

\begin{figure*}[t]
\centering
\noindent\makebox[0.9\textwidth][c]{\includegraphics[scale=0.5]{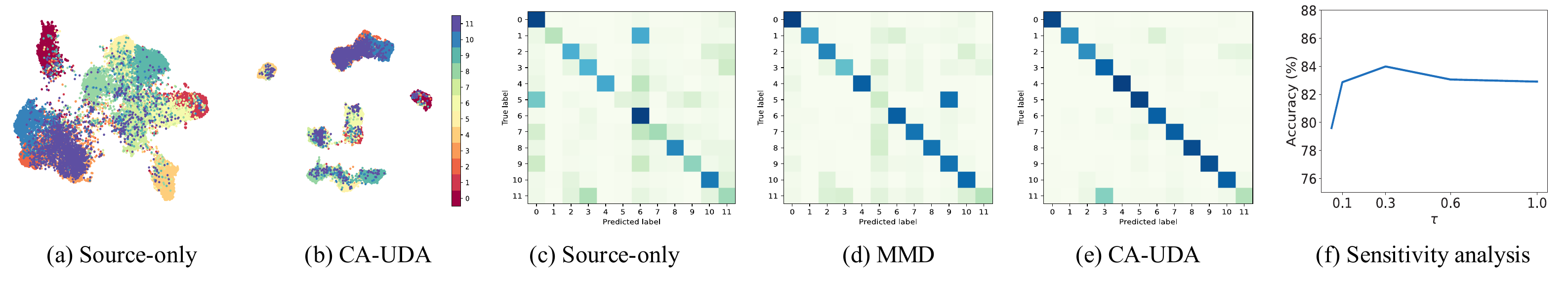}}
\caption{The visualizations of features (a-b) and confusion matrix (c-e) on VisDA-2017 from Source-only, MMD and CA-UDA with $\mathcal{L}_{sp}$. 0 to 11 correspond to the order of all categories listed in Table \ref{tab: visdacat}. (f) The sensitivity analysis of CA-UDA to $\tau$. (ResNet-50)} %\vspace{-3mm}
\label{fig:sensitivity}
\end{figure*}

%------------------------------------------------------------------------
\subsection{Ablation Studies}
\paragraph{Effect of the pseudo-label refinement.} To study the role of pseudo-label refinement in our CA-UDA, we remove the auxiliary target classifier (without $\mathcal{L}_{sp}$) as comparison. Table \ref{tab: digit}, \ref{tab: office}, \ref{tab: clef}, \ref{tab: visda}, \ref{tab: visdacat} show classification accuracy on all four benchmarks. Ours (with $\mathcal{L}_{sp}$) achieves more impressive improvements, which can be attributed to: 1) the easy-to-hard pseudo-label refinement step by the target-specific classifier, and 2) the confidence-check step by selecting more reliable samples for class-level alignment. It suggests effective pseudo-label refinement can provide substantial benefits to the class-aware domain alignment.
% As a result, falsely labeled target samples can be refined through target-specific classifier, and those hard examples that lies near margins can be removed from the training set.

% It proves that doing refinement on pseudo labels is an essential step in UDA, but it is commonly ignored in previous works.
% , i.e., pseudo-label refinement part with 
% the confidence check strategy
%------------------------------------------------------------------------

\begin{table}
\renewcommand\tabcolsep{4pt}
\begin{center}
\scalebox{0.7}{
\begin{tabular}{c|cccccc}
\hline
Dataset & $\mathcal{L}_{CE}^{t,pl_h}$& $\mathcal{L}_{CE}^{t,pl_s}$& $\mathcal{L}_{P2P}$& $\mathcal{L}_{C2C}$ & $\mathcal{L}_{C2C}$+$\mathcal{L}_1~\cite{16MCD}$ & $\mathcal{L}_{DA}$ \\
\hline
% VisDA-2017 &-& - & 73.48 & 74.22 & \textbf{85.45}\\
Office-31 & 82.46& 84.85 & 88.15 & 88.27 & 89.07 & \textbf{90.20} \\
ImageCLEF-DA & 87.42& 87.36 & 89.97 & 89.83 & 89.19 & \textbf{90.36} \\
\hline
\end{tabular}
}
\end{center}
\caption{Domain alignment evaluation with mean accuracy (\%).}
\label{tab: domain_alignment}
\end{table}

%------------------------------------------------------------------------

\vspace{-3mm}
\paragraph{Domain alignment evaluation.}
% Our class-level domain alignment $\mathcal{L}_{DA}$ consists of two components: $L_{C2C}$ in feature space and $L_{P2P}$ in label output space.
Table~\ref{tab: domain_alignment} evaluates our class-level domain alignment $\mathcal{L}_{DA}$ and its two components, i.e. $\mathcal{L}_{C2C}$ and $\mathcal{L}_{P2P}$. 
The performance is observed to drop when either one is removed. ``$\mathcal{L}_{CE}^{t,pl_h}$" and ``$\mathcal{L}_{CE}^{t,pl_s}$" denote two replacements of $\mathcal{L}_{DA}$, i.e. direct CE loss on pseudo-labels with hard one-hot and soft entropy weights~\cite{42labelpropagation}, respectively.
% Each sample in $\mathcal{L}_{CE}^{t,pl_h}$ is commonly assigned with one-hot weights, while in $\mathcal{L}_{CE}^{t,pl_s}$ is weighted by label entropy as in~\cite{42labelpropagation}. 
% the weight for each sample in CE loss is associated with the entropy to reflect the certainty of its pseudo label. 
In contrast, our $\mathcal{L}_{P2P}$ and $\mathcal{L}_{C2C}$ based settings greatly outperform CE-based supervisions by using pseudo labels for discrepancy minimization. 

To further evaluate $L_{P2P}$, an additional setting is considered, i.e. 
% $L_{P2P}$ computed in output space is evaluated by comparisons with two settings: 1) ``$\mathcal{L}_{C2C}$" removing $\mathcal{L}_{P2P}$ and 2)
``$\mathcal{L}_{C2C}+\mathcal{L}_1$" by replacing $\mathcal{L}_{P2P}$ with the L1-distance $\mathcal{L}_1$ as in~\cite{16MCD}. 
In Table~\ref{tab: domain_alignment}, $\mathcal{L}_1$ improves slightly over ``$L_{C2C}$" but drops on ImageCLEF-DA.
% which indicates
% $\mathcal{L}_1$ is sensitive to noisy pseudo-labels.
% which results in performance fluctuation over different datasets. 
In comparison, our $\mathcal{L}_{P2P}$ consistently improves the accuracy, particularly by a large margin on VisDA-2017. It implies that our $\mathcal{L}_{P2P}$ is effective on different datasets even with complex data diversity.
% in each calss. the stronger constraints on the distance of empirical probability means with the same class across domains. 
% Moreover, $\mathcal{L}_{P2P}$ performs particularly much better than the other two settings on the largest dataset VisDA-2017 (e.g., + 11.97\% over ``w/o. P2P"). It is because that this dataset is more diverse which results in the ambiguous boundaries among clusters. Whereas, under supervision of our $\mathcal{L}_{P2P} $, the large variances in each clusters are significantly minimized and distance among inter-clusters are forced to be larger, thus a more clear decision boundary can be obtained after optimization.

%------------------------------------------------------------------------

\vspace{-3mm}
\paragraph{Effect of the confidence-check strategy.}
% Table \ref{tab: abla_C} and \ref{tab: abla_CF_clef} 
Table \ref{tab: abla_pse_gen} examines our confidence-check strategy on two datasets, i.e. Office-31 and ImageCLEF-DA. ``w/o. CF" means we remove the confidence check and directly use the entire pseudo-labeled target data for domain alignment. It can be observed that overall results of ``w/o. CF" are inferior to ``w. CF", which demonstrates %the design of this component is reasonable by plugging the 
effectiveness of the confidence-check after the pseudo-label refinement step.
The higher accuracy of ``w. CF" is due to the reduction of detrimental influence from wrong pseudo-labels as hard examples with low confidence are filtered out.
% As target samples near the shared decision margins are more likely to be corrupted with false labels, hard examples with low confidence can be filtered out by checking the confidence through the selection condition. Therefore, class-level misalignment can be greatly alleviated.

%------------------------------------------------------------------------

\vspace{-3mm}
\paragraph{Pseudo-label generation evaluation.} %Clustering evaluation on target samples.
Table \ref{tab: abla_pse_gen} evaluates the quality of pseudo-labels generated by different approaches in our UDA task: 1) ``Net" by predictions from the network, 2) ``GMM" by estimations from the Gaussian Mixture Model, 3) ``Ours(OA)" by our cluster-based optimal assignment, and ``Ours(w/o. CF)" by additionally using pseudo-label refinement without data selection for alignment. All comparisons are conducted under the same setting that adopts $\mathcal{L}_{DA}$ on the entire data for domain alignment on Office-31 and ImageCLEF-DA.
% In addition, by considering the holistic model as the pseudo-labling CAT model~\cite{28CondDist2}, which employ a teacher model to provide pseudo-labels, is also included for evaluation. 
It is observed that our model can generate better pseudo labels by using optimal assignment and our pseudo-label refinement by training a task-specific network is effective in suppressing wrong pseudo-labeled samples for better domain alignment.

\begin{table}
\renewcommand\tabcolsep{4pt}
\begin{center}
\scalebox{0.65}{
\begin{tabular}{c|cccccc}
\hline
Dataset & Net & GMM & Ours(OA)&Ours (w/o. CF) & Ours (w. CF) \\
\hline
% VisDA-2017 & 73.48 & 74.22 & \textbf{85.45}\\
Office-31 & 88.47 & 89.68 & 90.20 &92.26& \textbf{94.01}\\
ImageCLEF-DA & 89.53 & 89.92 & 90.36 &91.30&\textbf{93.00}\\
\hline
\end{tabular}
}
\end{center}
\caption{Confidence-check and Pseudo-label generation evaluation with mean accuracy (\%).}
\label{tab: abla_pse_gen}
\vspace{-3mm}
\end{table}

% %------------------------------------------------------------------------

% %------------------------------------------------------------------------
% \begin{figure}[t]
% \centering
% \noindent\makebox[0.45\textwidth][c]{\includegraphics[scale=0.3]{iccv2021_CAUDA/LaTeX/figures/sensitity.eps}}
% \caption{The sensitivity analysis of CA-UDA to $\tau_1$ and $\tau_2$ with $\tau_1=\tau_2$ on VisDA-2017 (ResNet-50).} %\vspace{-3mm}
% \label{fig:sensitivity}
% \end{figure}

%------------------------------------------------------------------------
\subsection{Qualitative Results and Analysis}

\paragraph{Visualization.}
Figure \ref{fig:sensitivity} (a-e) shows visualizations of different models: 1) Source-only, 2) global alignment based MMD and 3) CA-UDA with $\mathcal{L}_{sp}$ on the VisDA-2017 dataset.
% We adopt ResNet-50 as the backbone.
Visualization of target features are done by the umap \cite{43umap}. Compared with MMD, our CA-UDA shows more compact intra-clusters, more apparent inter-cluster margins, and also less wrong labels. 
As visualized 
by the confusion matrix, a majority of the samples in some categories exhibit confusions for Source-only and MMD, while our CA-UDA significantly alleviates the class conditional shift.

\vspace{-3mm}
\paragraph{Sensitivity analysis.}
Figure~\ref{fig:sensitivity} (f) studies the sensitivity of $\tau_1$ and $\tau_2$ in Eq.~\ref{eq:DALoss} on VisDA-2017, which are balance weights of $\mathcal{L}_{C2C}$ and $\mathcal{L}_{P2P}$. For general analysis, we set $\tau_1=\tau_2$ and change the values over the range $\{0.05, 0.1, 0.3, 0.6, 1.0\}$. It can be observed that the accuracy curve is bell-shaped, where the performance steadily increases and then starts decreasing from 0.3. We set the best performing $\tau_1=\tau_2=0.3$ in all experiments.
% which is slightly fine-tuned around 0.3 when dealing with different settings and datasets.

%------------------------------------------------------------------------

\section{Conclusion}
 In this paper, we propose CA-UDA to improve the quality of the pseudo-labels and UDA results with optimal assignment, a pseudo-label refinement strategy and class-aware  domain alignment. In the pseudo-label refinement strategy, we show that the source domain bias in pseudo-label generation can be mitigated with the use of an auxiliary network trained on the target domain data. We further demonstrate that our optimal assignment can optimally align features in the source-to-target domains and our class-aware domain alignment can simultaneously close the domain gap while preserving the classification decision boundaries. Extensive experiments on several bench-mark datasets show that our method can achieve state-of-the-art performance in the image classification tasks.
 
% In this paper, we propose CA-UDA: a class-aware unsupervised domain adaptation method to mitigate the misalignment problem caused by underutilization of categorical information across the domains with optimal assignment and a pseudo-label refinement strategy. Under the class-based data sampling strategy and three-stage alternative optimization procedure, our method outperforms other existing methods on 4 benchmarks datasets.

{\small
\bibliographystyle{ieee_fullname}
\bibliography{egbib}
}

\end{document}